\documentclass[letterpaper, 10 pt, conference]{ieeeconf}  %

\IEEEoverridecommandlockouts                              %

\usepackage{xcolor}
\usepackage{url}
\usepackage{xfrac}
\usepackage{hyperref}
\usepackage{graphicx}
\usepackage{amsmath}
\usepackage{amssymb}
\usepackage{tabularx}
\usepackage{gensymb}
\usepackage{multirow}
\usepackage{multicol}

\usepackage{subcaption}
\usepackage{tabularray}
\usepackage{boldline}

\definecolor{LinkColor}{HTML}{105FC1}
\usepackage{hyperref}
\hypersetup{
    colorlinks=true,   
    linkcolor=black,
    filecolor=black, 
    urlcolor=LinkColor,
    pdftitle={MI-HGNN: Morphology-Informed Heterogeneous Graph Neural Network for Legged Robot Contact Perception},
    pdfpagemode=FullScreen,
}

\title{\LARGE \bf MI-HGNN: Morphology-Informed Heterogeneous Graph Neural Network for Legged Robot Contact Perception}
\author{Daniel Butterfield, Sandilya Sai Garimella, Nai-Jen Cheng, and Lu Gan%
\thanks{The authors are with the Georgia Institute of Technology, Atlanta, GA 30332, USA. Email: {\tt\small \{dbutterfield3, sgarimella34, ncheng49, lgan\}@gatech.edu.}}}

\begin{document}
\maketitle
\thispagestyle{empty}
\pagestyle{empty}

\begin{abstract} We present a Morphology-Informed Heterogeneous Graph Neural Network (MI-HGNN) for learning-based contact perception. The architecture and connectivity of the MI-HGNN are constructed from the robot morphology, in which nodes and edges are robot joints and links, respectively. By incorporating the morphology-informed constraints into a neural network, we improve a learning-based approach using model-based knowledge. We apply the proposed MI-HGNN to two contact perception problems, and conduct extensive experiments using both real-world and simulated data collected using two quadruped robots. Our experiments demonstrate the superiority of our method in terms of effectiveness, generalization ability, model efficiency, and sample efficiency. Our MI-HGNN improved the performance of a state-of-the-art model that leverages robot morphological symmetry by $\boldsymbol{8.4\%}$ with only $\boldsymbol{0.21\%}$ of its parameters. Although MI-HGNN is applied to contact perception problems for legged robots in this work, it can be seamlessly applied to other types of multi-body dynamical systems and has the potential to improve other robot learning frameworks. Our code is made publicly available at \href{https://github.com/lunarlab-gatech/Morphology-Informed-HGNN}{https://github.com/lunarlab-gatech/Morphology-Informed-HGNN}.

\end{abstract}

\section{Introduction}

Legged robots have significant advantages in navigating complex and unstructured environments. Their ability to traverse rough terrain and climb stairs and obstacles makes them an ideal platform for applications ranging from agriculture~\cite{garimella2021dandelion}, search and rescue~\cite{lindqvist2022multimodality}, home assistance~\cite{biswal2021development}, and planetary exploration~\cite{arm2023scientific}. For legged robots, contact perception is a problem using onboard sensor measurements to infer critical contact information such as contact state and forces~\cite{fourmy2021contact, 7397881}. Robust contact state classification can aid legged robot state estimation when walking on various terrains~\cite{hartley2018legged, hartley2018hybrid}, while accurately estimating external force helps robotic systems maintain balance, ensure stability, and adaptively adjust locomotion strategies in real-time~\cite{9636393, 5650416}.

As legged robots primarily interact with the environment through their feet, estimating the current foot contact state and ground reaction forces (GRFs) are important parts of contact perception. Accurately estimated GRF can directly influence the robot's ability to walk efficiently, prevent slipping, and adapt to changing terrains, ultimately improving both safety and performance in dynamic environments~\cite{cheng2023practice}.

Built-in contact sensors on a robot's feet can be used to directly detect contact state or measure GRFs. However, these sensors can be expensive and noisy~\cite{azimi2018robust}, prone to breakage, and change the weight distribution of the leg~\cite{hwangbo2016probabilistic}. Therefore, many robotic platforms lack built-in contact sensors, and estimated contact information is essential. A variety of estimation methods, including filtering-based~\cite{fakoorian2017ground,camurri_2017_IEEERAL}, smoothing-based~\cite{kang_2023_ICRA}, and observer-based~\cite{azimi2018robust} GRF estimators, have been developed over the years and exhibit satisfying performance for mechanical systems such as prosthetic legs. However, for complex dynamical systems operating in unknown physical environments, these approaches can be sensitive to sensor noises, unmodeled dynamics, and unforeseen disturbances~\cite{azimi2018robust}.

To mitigate these issues, more recently, learning-based approaches have been developed for contact perception problems~\cite{lin_2021_CORL, an2023artificial, arena2022ground}, demonstrating high accuracy and robustness across various environments. However, their performance is limited by the availability of large training datasets. In this work, we develop a learning-based contact perception framework with significantly improved generalization capability, model efficiency and sample efficiency, by constraining the learning problem using robot morphology.

\begin{figure}[t!]
    \centering
   \includegraphics[width=0.6\linewidth]{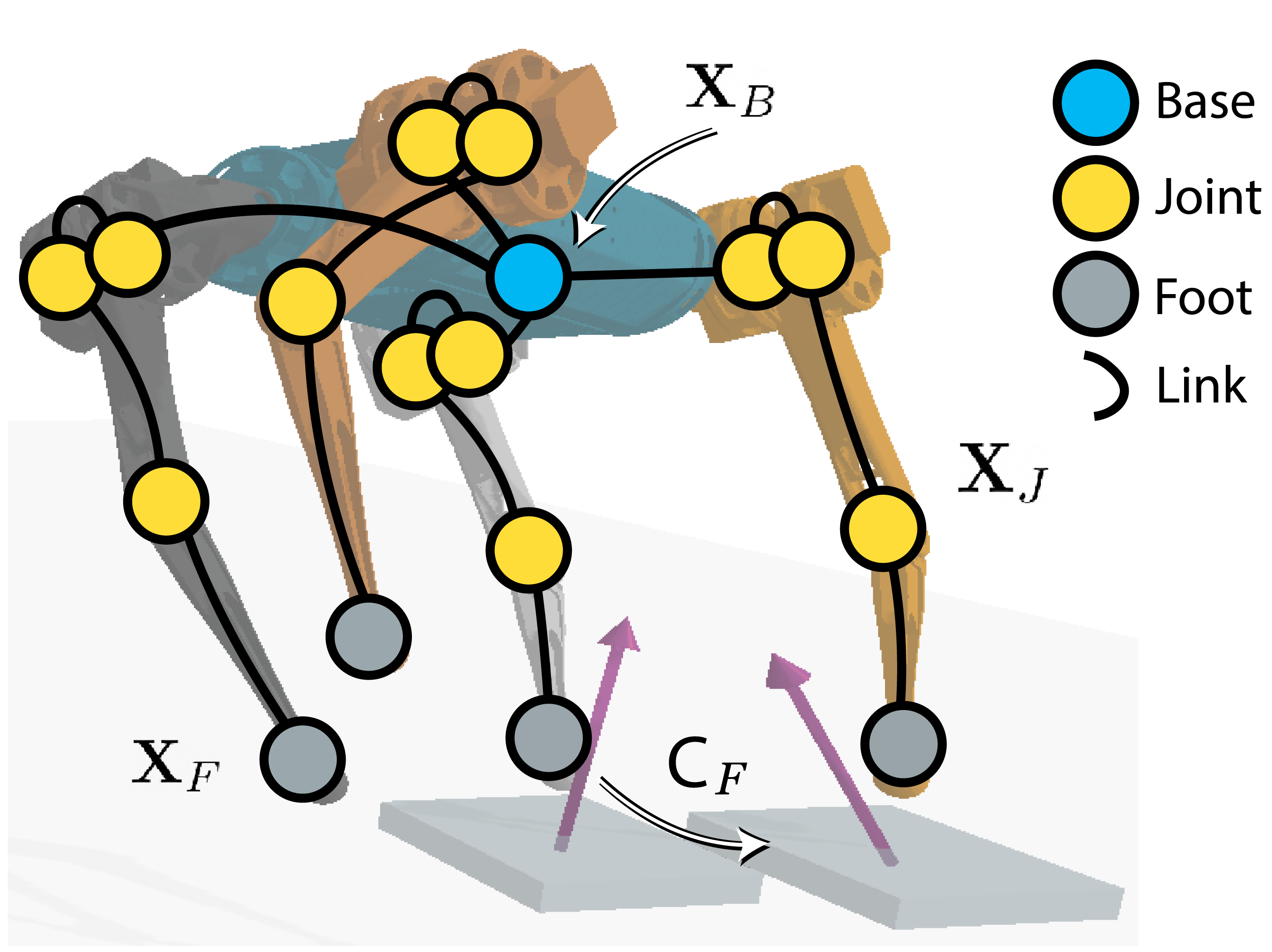}
    \caption{Visualization of our MI-HGNN for the Mini-Cheetah robot as an example. The structure and connectivity of our graph is constructed from the robot morphology. Local sensor measurements are embedded into the corresponding node to predict contact information at the foot node.}
    \label{fig:1}
\end{figure}

Our main contributions are:
\textbf{1.} We propose a novel neural network architecture, Morphology-Informed Heterogeneous Graph Neural Network (MI-HGNN), that can impose constraints in accord with the robot morphology to the learning problem. \textbf{2.} We develop a general and flexible learning-based contact perception framework for legged robots using the proposed MI-HGNN. \textbf{3.} We conduct extensive experiments using both real-world and simulated data, and demonstrate the superiority of our model in terms of effectiveness, generalization ability, model efficiency, and sample efficiency. \textbf{4.}~We make our code and data publicly available to foster the development of morphology-informed robot learning.

\section{RELATED WORK}
\label{related}

Legged robot contact perception refers to the ability of legged robots to sense and understand interactions between their feet and the ground or other surfaces. This involves detecting when and where contact occurs, the nature of the surface (e.g., hardness, texture), and the forces involved in the interaction. Contact perception is crucial for legged robots to maintain balance, adapt to uneven or unpredictable terrain, and execute tasks like walking and running.

\textbf{Model-based contact perception:}
Model-based contact perception involves using physics-based models~\cite{featherstone2014rigid} to predict and interpret the interaction between a robot and its environment, particularly during contact events. This approach uses a predefined model of the robot's dynamics, including its morphology and surface interaction, to estimate contact forces. For legged robots, some models assume certain terms in the equations of motion to be negligible~\cite{camurri_2017_IEEERAL, camurri_2020_FRA}, or only focus on the rows corresponding to the desired leg~\cite{kang_2023_ICRA}. Probabilistic methods, such as Markov models and Kalman filters, are also used to estimate the contact state~\cite{hwangbo2016probabilistic, bledt2018icra}. 
For GRF estimation, filtering-based~\cite{fakoorian2017ground, camurri_2017_IEEERAL}, smoothing-based~\cite{kang_2023_ICRA}, and observer-based~\cite{azimi2018robust} estimators have been developed.
While these models perform effectively under ideal conditions, they struggle to estimate contact forces in highly dynamic and complex scenarios accurately.

\textbf{Learning-based contact perception:}
Learning-based contact perception involves learning techniques for robots to learn contact events through data-driven models. These models are trained on datasets of interactions between the robot and its environment, allowing the robot to adapt to complex and dynamic conditions. Learning-based contact perception is beneficial in scenarios where model-based approaches struggle, such as irregular terrains or unpredictable environments, as it allows the robot to improve the estimation of GRFs~\cite{arena2022ground,an2023artificial,albadin2025estimation} or contact state~\cite{lin_2021_CORL}. Learning-based methods face challenges due to their need for large labeled datasets, which are costly to collect in dynamic environments. They also function as ``black-box" models, making them hard to interpret compared to physically grounded approaches. Additionally, they may ignore physical constraints, leading to unrealistic or unsafe behavior.

\textbf{Geometric deep learning for robots:} Instead of relying solely on the data to learn a model, geometric deep learning incorporates physical, geometric, or morphological constraints into the learning problem to make the most of a small sample~\cite{zhang_2022_RAL,sferrazza2024body,huang2020one}. Ordonez et al. propose a theoretical framework for identifying discrete morphological symmetries (DMSs) of a dynamical system as a symmetry group $\mathcal{G}$, which can be exploited for augmenting robotic datasets or constraining learning models~\cite{ordonez2023discrete}. A subset of geometric deep learning models, graph neural networks~(GNNs), can model relational, temporal, or spatial relationships between nodes and exploit these relationships to learn complex functions encountered in \mbox{robotics~\cite{li_2020_IROS,kim_2021_ARXIV,sanchez2020learning,almeida_2021_IROS}}. Similar to our method, WAGNN~\cite{zhang_2020_RAS} and Sanchez-Gonzalez et al.~\cite{sanchez2018graph} both construct GNN architectures from the robot morphology, but employ homogeneous graphs with no ability to extract distinct features for different node types. \mbox{NerveNet~\cite{wang_2018_ICLR}} also leverages a GNN derived from the kinematic tree for reinforcement learning, but differs from our method by casting both links and joints into nodes, resulting in a less compact graph compared to ours. Inspired by the morphological symmetries found in legged robots, our \mbox{MI-HGNN} imposes shared weights between morphologically-identical kinematic branches to improve model generalization and sample efficiency.

\section{Preliminaries}

\subsection{Legged Robot Dynamics}

\begin{figure*}[ht] 
    \centering
    \includegraphics[width=\textwidth]{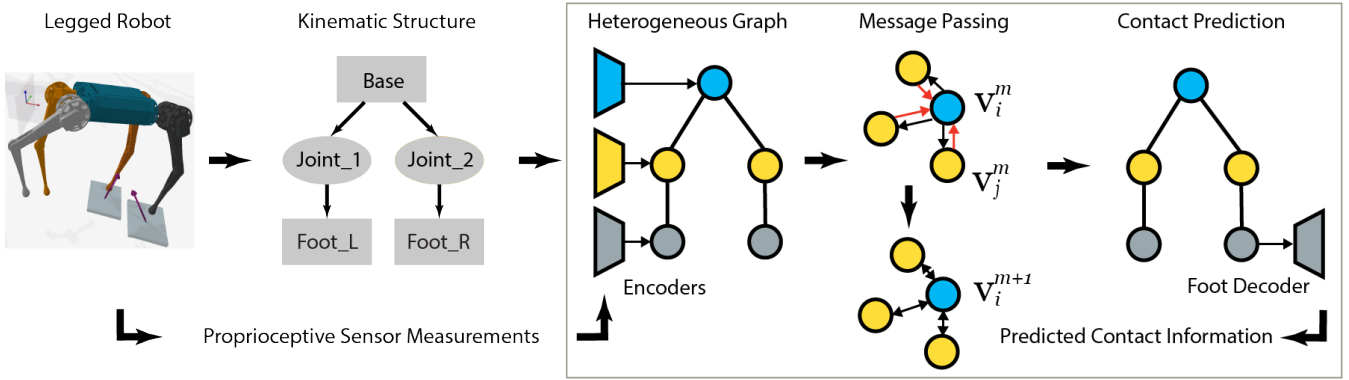}  
    \caption{Overview of the proposed MI-HGNN for legged robot contact perception problems. Our MI-HGNN is constructed from a robot kinematic structure where nodes are joints and edges are links. Proprioceptive sensor measurements acquired at each local frame are embedded into the corresponding node through a heterogeneous encoder, and fused via several layers of Message Passing. A foot decoder attached to the foot node exacts the contact information during inference.}
    \label{fig:pipeline}
\end{figure*}

Legged robots such as quadrupeds and humanoids are typically treated as floating-base multi-body systems and are described by $n_b$ unactuated base coordinates $(\boldsymbol{q}_b)$ and $n_j$ actuated joint coordinates \mbox{$(\boldsymbol{q}_j)$; as $\boldsymbol{q}=\begin{bmatrix}\boldsymbol{q}_b, \boldsymbol{q}_j \end{bmatrix}^T$}. In these generalized coordinates, $\boldsymbol{q}_b \in SE(3)$ represents the translation and orientation of the floating base $\mathcal{B}$ w.r.t. the inertial frame~$\mathcal{I}$. Each value in $\boldsymbol{q}_{j} \in \mathbb{R}^{n_j}$, as $\boldsymbol{q}_{j, i}$, corresponds to the 1 DoF of the $i$th joint $\mathcal{J}_i$. The generalized velocity, acceleration, and torques of the system are $\dot{\boldsymbol{q}}$, $\ddot{\boldsymbol{q}}$, and $\boldsymbol{\tau}$, respectively. The dynamics for this floating-base system are:
\begin{equation} \label{eq:dynamics}
            \boldsymbol{M}(\boldsymbol{q})\ddot{\boldsymbol{q}} + \boldsymbol{C}(\boldsymbol{q},\dot{\boldsymbol{q}}) + \boldsymbol{g}(\boldsymbol{q}) = \boldsymbol{S}^T\boldsymbol{\tau} + {\boldsymbol{J}_{\textit{ext}}({\boldsymbol{q}})}^T \boldsymbol{F}_{\textit{ext}}
\end{equation}
where $\boldsymbol{M}(\boldsymbol{q})$ is the mass matrix, $\boldsymbol{C}(\boldsymbol{q},\dot{\boldsymbol{q}})$ are the coriolis and centrifugal terms, $\boldsymbol{g}(\boldsymbol{q})$ are the gravity terms, $\boldsymbol{S}^T$ is the selection matrix of actuated joints, $\boldsymbol{F}_{\textit{ext}}$ are the external forces, and ${\boldsymbol{J}_{\textit{ext}}({\boldsymbol{q}})}$ are the external force contact \mbox{Jacobians~\cite{kang_2023_ICRA,featherstone2014rigid}}.

\subsection{Model-Based Contact Force}

Given proprioceptive information ($\boldsymbol{q}$, $\dot{\boldsymbol{q}}$, $\ddot{\boldsymbol{q}}$, $\boldsymbol{\tau}$), we want to solve for the contact forces. Assuming that $\boldsymbol{F}_{\textit{ext}}$ only comprises of the leg contact forces for $n_l$ legs (ignoring wind and other disturbances), and denoting this value as $\boldsymbol{F}_{\textit{c}}$ (where $\boldsymbol{F}_{\textit{c}} = [\boldsymbol{F}^T_1, ..., \boldsymbol{F}^T_{n_l}]^T \in \mathbb{R}^{3n_l}, \boldsymbol{F}_l = [x, y, z]^T$, and $\boldsymbol{J}_l$ is the external force contact Jacobian for leg $l$), we can rewrite Eq.~\ref{eq:dynamics} as:
\begin{equation} \label{eq:dynamics_force}
    \boldsymbol{M}(\boldsymbol{q})\ddot{\boldsymbol{q}} + \boldsymbol{C}(\boldsymbol{q},\dot{\boldsymbol{q}}) + \boldsymbol{g}(\boldsymbol{q}) = \boldsymbol{S}^T\boldsymbol{\tau} + \sum_{l = 1}^{n_l}{\boldsymbol{J}_l({\boldsymbol{q}})}^T \boldsymbol{F}_l
\end{equation}

Assuming the effect of the other legs to be negligible for calculating $\boldsymbol{F}_{\textit{l}}$ for a specified leg, 
Eq.~\ref{eq:dynamics_force} can be separated into $\boldsymbol{n}_l$ different equations; one for each leg:
\begin{equation} \label{eq:force}
     \boldsymbol{F}_{\textit{l}} = -({\boldsymbol{J}_{\textit{l}}({\boldsymbol{q}_l})}^T)^{-1}(\boldsymbol{S}^T_l\boldsymbol{\tau}_l - \boldsymbol{M}_l(\boldsymbol{q}_l)\ddot{\boldsymbol{q}_l} - \boldsymbol{C}_l(\boldsymbol{q}_l,\dot{\boldsymbol{q}_l}) - \boldsymbol{g}_l(\boldsymbol{q}_l)) 
\end{equation}
where $\boldsymbol{q}_l$ contains the base coordinates $\boldsymbol{q}_b$ and a subset of the joint coordinates $\boldsymbol{q}_j$ corresponding to the given leg $l$.

This floating-base model has been used in many previous works for contact state estimation or GRF calculation on a quadruped with slightly modified \mbox{assumptions~\cite{camurri_2017_IEEERAL,kang_2023_ICRA,camurri_2020_FRA}}. Based on these formulations, each of the legs on a legged robot share a similar underlying function for contact force calculation. This relationship inspires us to share our MI-HGNN model parameters amongst the system's legs to learn a generalizable dynamics equation similar to Eq.~\ref{eq:force}.

\section{Morphology-Informed Neural Network}

In this section, we introduce the proposed MI-HGNN and how it is used for legged robot contact perception problems. We first construct a heterogeneous graph based on robot morphology, where each node in the graph corresponds to a local coordinate frame in a robot's kinematic structure. We then fuse multi-modal sensor measurements acquired in different local frames via message passing. The contact information we want to predict can thus be extracted from the corresponding nodes in the final graph.

\subsection{Robot Morphology as a Heterogeneous Graph}
\label{sec:graph}

We represent the morphology of an legged robot as a heterogeneous graph~\cite{zhang2019heterogeneous}, denoted as $\mathcal{G}=(\mathcal{V}, \mathcal{E})$, consisting of a node set $\mathcal{V}$ associated with a node type mapping: \mbox{$\phi: \mathcal{V} \rightarrow \mathcal{A}$}, and an edge set $\mathcal{E}$ associated with an edge type mapping: \mbox{$\psi: \mathcal{E} \rightarrow \mathcal{R}$}. Considering a legged robot morphology that has a base body and limbs connected via joints, we define three types of nodes in the heterogeneous graph: \mbox{$A \in \mathcal{A} = \{B, J, F\}$}, including robot \emph{base} (B), \emph{joint} (J), and \emph{foot} (F), respectively. An edge type is therefore defined according to the type of nodes the edge connects to: \mbox{$\psi(e_{i,j}) = \langle \phi(v_i), \phi(v_j) \rangle$}, e.g., $\langle B, J \rangle$ and $\langle J, F \rangle$.

We construct this heterogeneous graph directly from a robot's kinematic structure; forming the node set $\mathcal{V}$ from the joints and the edge set $\mathcal{E}$ from the links. We note that the robot's base body and feet are fixed joints whereas others are all revolute joints with an actuator, which is one factor that inspires us to design a heterogeneous graph. After construction, the connectivity of this heterogeneous graph is in accord with the specific robot morphology, which makes our network and learning algorithm morphology-aware.

\subsection{Sensor Fusion as Message Passing}
\label{sec:message_pass}

We formulate the learning-based contact perception problem as to learn a function that maps multi-modal proprioceptive measurements to contact information such as binary contact states or continuous GRFs for all robot feet: \mbox{$ \Theta : \mathcal{X} \rightarrow \mathcal{Y} \in \mathbb{R}^{l \times n}$}, where $l$ is the number of feet and $n$ is the dimension of the predicted contact information. Given the \emph{distributed} sensor measurements in each local frame \mbox{$\boldsymbol{x}=\{\boldsymbol{x}_{i}\}$}, we assign the local sensor data as the input vector to the corresponding node $v_i$ in our heterogeneous graph, such that a node of type \mbox{$\phi(v_i)= A$} only receives sensor measurements specific to that node type: $\boldsymbol{x}_A$. For instance, a joint node will only receive joint encoder measurements. The heterogeneity also enables more flexible sensor fusion as nodes of different types can take inputs with various lengths and in different modalities.

Considering the aforementioned sensory multi-modality, we attach a heterogeneous encoder to each node $\boldsymbol{E}_A, A\in \mathcal{A}$ to learn embeddings specific to a node type and fuse sensor measurements via a graph neural network's message passing. Overall, the proposed MI-HGNN architecture is constructed from the heterogeneous encoders $\boldsymbol{E}_A$, a number of Message-Passing layers \mbox{$\boldsymbol{M}^m, m=\{0, ..., n_m\}$}, and a decoder $\boldsymbol{D}_{F}$ applied to the foot nodes, with $\boldsymbol{\varphi}$ as a non-linear activation function:
\begin{equation}
\begin{aligned}
     \boldsymbol{v}_{i}^0 & = \boldsymbol{\varphi}(\boldsymbol{E}_{\phi(v_i)}(\boldsymbol{x}_i)), \quad v_i \in \mathcal{V}\\
     \boldsymbol{v}^{m+1} & = \boldsymbol{\varphi}(\boldsymbol{M}^m(\boldsymbol{v}^{m}, \mathcal{E})),
    \quad \text{while $m < n_m$} \\
    \boldsymbol{y}_{i} & =\boldsymbol{D}_F(\boldsymbol{v}^{n_m}_{i}), \quad \text{for $\phi(v_i) = F$}
\end{aligned}
\end{equation}
with the following Message-Passing formulation:
\begin{equation}
\boldsymbol{M}^m(\boldsymbol{v}_i, \mathcal{E}) = \sum_{A \in \mathcal{A}} (\boldsymbol{W}_{\mathcal{R}, 1}\boldsymbol{v}_i  + \boldsymbol{W}_{\mathcal{R}, 2}(\!\!\sum_{j \in N_A(i)}\!\!\!\!\boldsymbol{v}_j))
\end{equation}
where $\boldsymbol{v}_i$ is the embeddings of node $v_i$, $N_A(i)$ is the set of neighbor nodes connected to node $v_i$ of type $A$, and $\boldsymbol{W}$ is a learned matrix that depends on the current edge type $\mathcal{R} = \langle A, \phi(v_i) \rangle $. We use a form of the graph operator from~\cite{morris2019weisfeiler} that is modified for use with heterogeneous message passing. Our heterogeneous encoders $\boldsymbol{E}_A$ and edge-informed aggregation strategy can enable more effective multi-modal sensor fusion than their homogeneous counterparts used in~\cite{wang_2018_ICLR}, as node features are computed with explicit modality awareness.

\subsection{Model Training}
\label{sec:loss}
Given ground truth labels ($\boldsymbol{c}$) at each foot node, our MI-HGNN is trained via supervised learning. Due to its modularity, our MI-HGNN directly predicts values on foot nodes. Let $n_f$ be the number of nodes $v$ where $\phi(v_i) = F$. Thus, our output is $\boldsymbol{y} \in \mathbb{R}^{n_f \times n_y}$, with $\boldsymbol{y}_{i} \in \mathbb{R}^{n_y}$ as the output of node $v_i$ with output size $n_y$, and $\boldsymbol{c} \in \mathbb{R}^{n_f}$ as the labels. For classification problems such as contact state prediction, our model outputs probability logits for each foot ($n_y$ = 2), leading us to use a foot-wise cross-entropy loss $\boldsymbol{L}(\boldsymbol{y}, \boldsymbol{c}) = \sum_i^{n_f} \boldsymbol{L}_{\text{CE}}(\boldsymbol{y_i}, c_i)$. For a regression problem such as GRF estimation, we directly output predicted values per foot node ($n_y$ = 1) and similarly use a foot-wise mean squared error loss $\boldsymbol{L}(\boldsymbol{y}, \boldsymbol{c}) = \sum_i^{n_f} \boldsymbol{L}_{\text{MSE}}(\boldsymbol{y_i}, c_i)$.

\subsection{Model Interpretation}
\label{sec:discussions}

\textbf{A data correlation aspect:} 
MI-HGNN constrains the learning problem based on the robot's morphology and explicitly captures the correlation and causality between inputs. This configuration mirrors the flow of information in a robotic system, as represented by a kinematic tree. Message passing between nodes is influenced by intermediate nodes, resembling the physical laws of the robot's system, thereby embedding causality into the message-passing process. As a result, our neural network achieves comparable performance to other methods with fewer parameters, as it does not need to learn the implicit relationships between data.

\textbf{A symmetry aspect:} As the Message-Passing layer $\boldsymbol{M}$ varies only according to the edge types $\mathcal{R}$, the learned weights $\boldsymbol{W}$ are identical between all morphologically-identical limbs of the robotic platform. In addition to reducing the complexity of the search space and lowering model sizes, these constraints help us learn a model that is shared by the four legs for quadruped robots. Although this over-constrains the learning problem, empirically we find that this weight sharing allows the model to quickly learn a more generalizable function when compared to completely unconstrained learning models, leading to better performance.

\section{EXPERIMENTS}
\label{sec:experiments}

In this section, we evaluate the proposed MI-HGNN on two common contact perception problems for legged robots: contact state detection (classification) and ground reaction force estimation (regression). The former experiment is conducted on a real-world dataset collected using an MIT Mini-Cheetah robot, while the latter is conducted in a simulation environment for an A1 robot.

\subsection{Contact Detection}

\textbf{Dataset:} The Mini-Cheetah contact dataset presented in~\cite{lin_2021_CORL} is collected using the robot operating in different gaits and on various terrains, including \emph{asphalt road}, \emph{concrete}, \emph{forest}, \emph{grass}, etc. It consists of synchronized proprioceptive sensor measurements and annotated contact state labels at 1000 Hz. The synchronized data at each time step includes joint angle, joint angular velocity from 12 joint encoders (\mbox{$\boldsymbol{q} \in \mathbb{R}^{12}$}, \mbox{$\dot{\boldsymbol{q}} \in \mathbb{R}^{12}$}), base linear acceleration, base angular velocity from IMU (\mbox{$\boldsymbol{a}_b \in \mathbb{R}^3$}, \mbox{$\boldsymbol{\omega}_b \in \mathbb{R}^3$}), each of the 4 foot positions and velocities obtained via forward kinematics (\mbox{$\boldsymbol{p}_l \in \mathbb{R}^{3}$}, \mbox{$\boldsymbol{v}_l \in \mathbb{R}^{3}$}, \mbox{$l=\{LF, LH, RF, RH\}$}), and a binary contact state for each leg (\mbox{$\boldsymbol{c} \in \mathbb{R}^4$}, \mbox{$c_l \in \{0,1\}$}, where 1 indicates a firm contact with the ground and 0 otherwise). We use the modified train/val/test data split introduced in~\cite{ordonez2023discrete} throughout this experiment, in which the test set includes recordings on unseen ground types and gait types, to evaluate the generalization ability of our model on unseen scenarios.

\textbf{Baselines:} We compare our model with three state-of-the-art learning-based contact detection models: the original convolutional neural network (CNN) proposed in~\cite{lin_2021_CORL}, the same CNN trained with augmented data by performing $\mathcal{C}_2$ symmetry transformations on the original training data (CNN-aug)~\cite{ordonez2023discrete}, and the $\mathcal{C}_2$-equivariant CNN that incorporates hard-equivariance constraints in the network (ECNN)~\cite{ordonez2023discrete}. We choose the CNN-aug and ECNN as our baselines due to their morphology awareness achieved by leveraging the Mini-Cheetah's morphological symmetry.

\textbf{Implementation:} For a fair comparison, we follow the experiment setup in \cite{lin_2021_CORL, ordonez2023discrete}. The input data for all baseline models is a history of 150 samples \mbox{$\boldsymbol{x}=\{\boldsymbol{z}_i\}_{i=0}^{150} \in \mathbb{R}^{54 \times 150}$}, \mbox{$\boldsymbol{z}=[\boldsymbol{q}, \dot{\boldsymbol{q}}, \boldsymbol{a}_b, \boldsymbol{\omega}_b, \boldsymbol{p}, \boldsymbol{v}] \in \mathbb{R}^{54}$}, and the output is a predicted contact state \mbox{$\boldsymbol{y} \in \mathbb{R}^{16}$} representing the logits of the 16 different contact state combinations, i.e., one binary contact state per leg. Due to the graph structure of our model, we re-organize the input data $\boldsymbol{x}$ based on the corresponding node index $j$ and node type $A$: \mbox{$\boldsymbol{x} = \{\boldsymbol{x}_{B}, \boldsymbol{x}_{J}, \boldsymbol{x}_{F}\}$}, where $\boldsymbol{x}_{B} = [\mathbb{F}({\{\boldsymbol{a}_b\}_{i=0}^{150}}$, ${\{\boldsymbol{\omega}_b\}_{i=0}^{150}})] \in \mathbb{R}^{1 \times 900}$, \mbox{$\boldsymbol{x}_{J} \in \mathbb{R}^{12 \times 300}$}, $\boldsymbol{x}_{J, j} =  [\mathbb{F}({\{\boldsymbol{q}_j\}_{i=0}^{150}}$, ${\{\dot{\boldsymbol{q}_j}\}_{i=0}^{150}})] \in \mathbb{R}^{300}$, $\boldsymbol{x}_{F}  \in \mathbb{R}^{4 \times 900}$, \mbox{$\boldsymbol{x}_{F, j} =  [\mathbb{F}({\{\boldsymbol{p}_l\}_{i=0}^{150}}$}, ${\{\boldsymbol{v}_l\}_{i=0}^{150}})] \in \mathbb{R}^{900}$, with $\mathbb{F}$ as a flatten operator. The output from our model is directly the logit of a binary contact state for each corresponding foot, allowing the usage of a foot-wise cross-entropy loss defined in~\ref{sec:loss}. We argue that this loss function leads to a formulation of a binary node classification problem, which is more natural for contact perception problems, compared to the 16-class classification formulation in~\cite{lin_2021_CORL, ordonez2023discrete}.

In this experiment, the baselines are trained using the original network architecture and learning rates provided in~\cite{ordonez2023discrete}. Our MI-HGNN is trained with a hidden size of 128, i.e., the length of the node embeddings, and 8 layers of message passing, with a learning rate of $10^{-4}$. All models are trained using a batch size of 30, use time-wise normalization per dataset entry, and employ an early-stopping mechanism monitoring validation loss.

\begin{figure}[t!]
    \centering
    \includegraphics[width=1.0\linewidth]{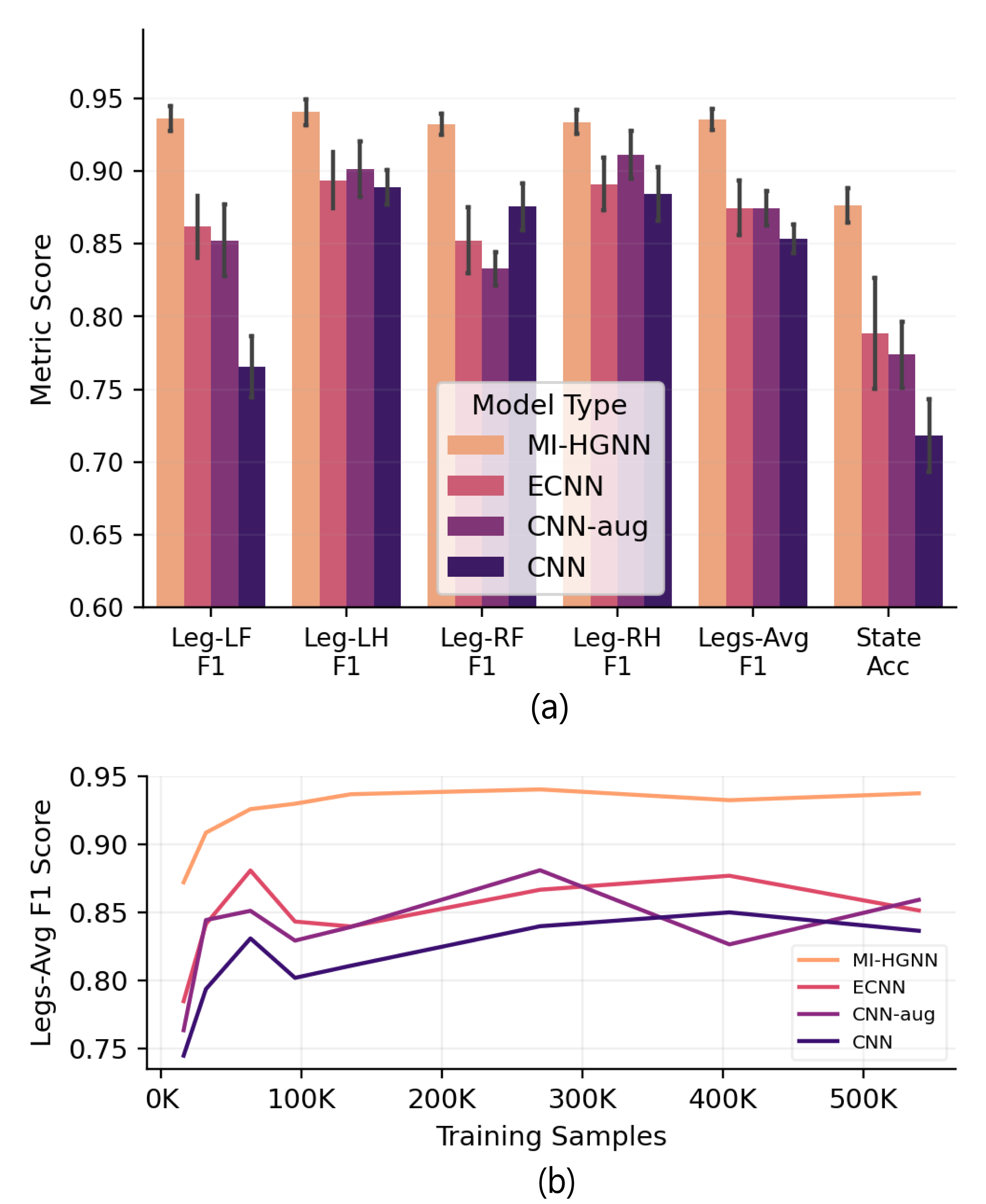}
    \caption{Contact detection results on the real-world Mini-Cheetah contact dataset~\cite{lin_2021_CORL}: (a) classification performance of four models on the unseen test set, trained with the entire training set. The mean and standard deviation across 8 runs are reported. (b) sample efficiency evaluation for all models.}
    \label{fig:contact_exp}
\end{figure}

\textbf{Results:} For a fair comparison, we evaluate all models using the same metrics used in~\cite{ordonez2023discrete}, namely foot-wise binary F1-score, averaged F1-score (the mean of the four leg F1-scores), and the contact state accuracy evaluated on 16 states, which is a harsher metric than averaged accuracy as the prediction counts as accurate only when the predicted state is correct for all 4 feet. To evaluate our model using the contact state accuracy metric, we convert our foot-wise prediction into a 16-state prediction for calculation. The results of our experiment are shown in Fig.~\ref{fig:contact_exp} (a).

From the results, our model outperforms the three baselines on all metrics by a significant margin. In terms of contact state accuracy, our MI-HGNN achieves a mean accuracy of $87.7\%$ across 8 runs, improving the performance of the second best model (i.e., ECNN with $78.8\%$ mean accuracy) by $11.3\%$. Compared with the \emph{morphology-agnostic} CNN model (with $71.8\%$ mean accuracy), our model outperforms it by $22.1\%$. As the evaluation is conducted on out-of-training-distribution data, our MI-HGNN model shows superior generalization capability for the contact detection problem compared to baselines. In addition, our model is more robust to dataset biases compared with other models: it performs equally well for the four legs despite the dataset's heavy bias to certain contact states~\cite{ordonez2023discrete}, whereas other models are affected by the data biases and favor certain legs. Our conjecture is that this is attributed to the shared weights across four legs in our MI-HGNN and the employment of the foot-wise entropy loss. It is worth mentioning that our model also exhibits the best reliability among the four models with the lowest standard deviation across multiple runs.

Besides the significant improvement in contact detection accuracy, our model also has the highest model efficiency. The parameter size of each model is reported in Table \ref{table:modelParams}. Our MI-HGNN primarily learns a set of small shared Message-Passing weight matrices \mbox{$\boldsymbol{W} \in \mathbb{R}^{h \times h}$} with a relatively small hidden size $h$. This leads to significantly reduced model size when compared with CNN or MLP-based models. As shown in Table~\ref{table:modelParams}, our MI-HGNN only contains a size of $28\%$ and $15\%$ of the parameters of the ECNN and CNN models, respectively, while achieving better performance as shown in Fig.~\ref{fig:contact_exp} (a). By sharing learned weights across different topologically identical limbs or kinematic branches of a robot, the MI-HGNN achieves higher accuracy simultaneously with higher computational efficiency.

\begin{table}
    \caption{Parameter size of each model in two experiments.}
    \label{table:modelParams}
    \centering
\begin{tabular}{l|l|>{\centering\arraybackslash}p{2.5cm}}
     \hline 
     Task & Model  & \# of Parameters \\
     \hline
     \multirow{3}{*}{\shortstack[l]{Contact Det. \\ (classification)}} & CNN  / CNN-aug  & 10,855,440 \\
       & ECNN  & 5,614,770   \\
     & MI-HGNN (ours) & 
     \textbf{1,585,282} \\
     \hline
     GRF Est. & MLP &  1,582,604   \\
     (Regression) & MI-HGNN (ours) & 
     \textbf{1,489,281} \\
     \hline
    \end{tabular}
\end{table}

As our MI-HGNN has high model efficiency, we are interested in studying its sample efficiency. In this experiment, we train the four models with various fractions of the entire training set and evaluate the legs averaged F1-score on the test set. The results are given in Fig.~\ref{fig:contact_exp} (b). Overall, we find that the performance of our model does not drop significantly until the number of training samples is reduced to $10\%$ of the entire training set. In addition, competitive results can still be achieved by our model using only $2.5\%$ of the training set, i.e., 15863 training samples.

\textbf{Ablation study:} Lastly, we conduct an ablation study on model selection by varying two important hyperparameters in our model design, i.e., the number of Message-Passing layers and the hidden size, leading to 12 different models in total. The resulting model size and test contact-state accuracy are reported in Table \ref{table:ablation}. From the results, we can see that our MI-HGNN is not sensitive to the hyperparameters. Using our smallest MI-HGNN model, we reduce the model parameters from ECNN by a factor of 482 and still improve its performance by $8.4\%$, when compared with the ECNN accuracy from Fig.~\ref{fig:contact_exp} (a).

\begin{table}[t!]
    \caption{Ablation study on model selection by varying two hyperparameters of our MI-HGNN.}
    \label{table:ablation}
    \centering
    \setlength\tabcolsep{2pt}
    \begin{tabular}{c|c|c|c}
    \hline
        \# of Layers &  Hidden Size & \# of Parameters & Test State Acc. \\
        \hline
        4  & 5   &    11,627   & 0.8543 \\
        4  & 10  &    25,252         & 0.87362 \\
        4  & 25  &    78,127         & \textbf{0.87666} \\
        4  & 50  &   206,252         & 0.85342 \\
        4  & 128 &   927,362         & 0.86033 \\
        4  & 256 & 3,165,442         & 0.85885 \\
        \hline
        8  & 50  &   307,252         & 0.85052 \\
        8  & 128 & 1,585,282         & 0.87953 \\
        8  & 256 & 5,792,002         & \textbf{0.89343} \\
        \hline
        12 & 50  &   408,252         & 0.87886 \\
        12 & 128 & 2,243,202         & 0.88332 \\
        12 & 256 & 8,418,562         & \textbf{0.90117} \\
    \hline
    \end{tabular}
\end{table}

\subsection{Ground Reaction Force Estimation}

\textbf{Dataset:}
To evaluate the performance of our model on another contact perception problem, i.e., GRF estimation, we generate our own simulated GRF dataset in this experiment. We record sensor data and the corresponding ground truth GRFs by operating an A1 robot in the Quad-SDK simulator~\cite{norby2020iros, norby2022icra, compHSL}. The generated dataset consists of synchronized proprioceptive sensor measurements at a maximum of 500 Hz, including joint angle, joint angular velocity, and joint torque from 12 joint encoders (\mbox{$\boldsymbol{q} \in \mathbb{R}^{12}$}, \mbox{$\dot{\boldsymbol{q}} \in \mathbb{R}^{12}$}, \mbox{$\boldsymbol{\tau} \in \mathbb{R}^{12}$}), base linear acceleration, base angular velocity from IMU (\mbox{$\boldsymbol{a}_b \in \mathbb{R}^3$}, \mbox{$\boldsymbol{\omega}_b \in \mathbb{R}^3$}), and GRFs for each leg in the Z direction (\mbox{$\boldsymbol{c} \in \mathbb{R}^4$}). It also includes the ground truth robot pose in the world frame, represented as a translation and a quaternion (\mbox{$\boldsymbol{T} \in \mathbb{R}^{3}$}, \mbox{$\boldsymbol{\omega} \in \mathbb{R}^{4}$}).

\begin{figure}
    \centering
   \includegraphics[width=1.\linewidth]{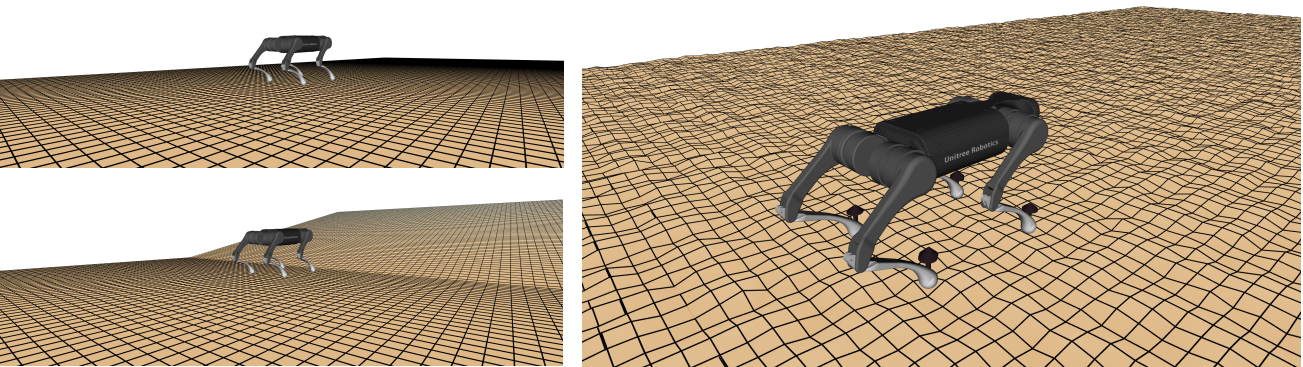}
    \caption{Examples of various terrains used for GRF data collection: top-left: flat terrain; bottom-left: 20$\degree$ slope; right: rough terrain. GRF for each leg in Z direction is visualized as the black arrow.}
    \label{fig:terrains}
\end{figure}

To increase the data variety, we collect 8 train/val sequences comprised of two terrain friction coefficients (\mbox{$\mu \in \{0.75,1.0\}$}), two terrain slopes (\textit{flat}, \textit{20$\degree$}), and two walking (in trot gait) speeds (\mbox{$\dot{x} \in \{0.5 \text{ m/s}, 0.75\text{ m/s}\}$}). For testing, we collect 13 test sequences that introduced \emph{unseen} parameters (\mbox{$\mu = 0.5, \dot{x} = 1.0 \text{ m/s}$}, and \textit{rough} terrain), including a sequence with these parameters all unseen (``Unseen All"), which are designed to mimic unknown environments for out-of-distribution testing. In total, our dataset comprises of 530,779 synchronized data samples. A visualization of the various data collection environments is shown in Fig.~\ref{fig:terrains}.

\begin{table}[t!]
    \caption{GRF estimation performance (RMSE) of three methods on simulated unseen test sequences.}
    \label{table:regression_table}
    \centering
    \setlength\tabcolsep{5pt}
    \begin{tabular}{p{1.8cm}|c | c |c }
    \hline
         Test Sequence &  MI-HGNN & MLP & FBD \\
        \hline
        Unseen Friction & \textbf{8.113} $\pm$ 0.09  & $\ \ 8.978 \pm 0.24$ & $29.156$  \\
        Unseen Speed    & $\ \ \textbf{9.788} \pm 0.11$ & $10.940\pm 0.15$ & $29.429$ \\
        Unseen Terrain  & $\ \ \textbf{8.804} \pm 0.12$ & $10.589 \pm 0.27$ & $29.195$ \\
        Unseen All      & $ \textbf{10.283} \pm 0.08$ & $12.466 \pm 0.45$ & $29.708$ \\
        \hline
        Unseen Total  & $\ \ \textbf{9.038} \pm 0.09$ & $10.455 \pm 0.23$ & $29.294$ \\

    \hline
    \end{tabular}
\end{table}

\textbf{Baselines:} We compare our model with two baselines; a fully-connected MLP~\cite{an2023artificial}, and the model-based contact force prediction using floating base dynamics (FBD) (Eq.~\ref{eq:dynamics_force}), implemented for A1 robot using Pinocchio~\cite{carpentier2019pinocchio}.

\textbf{Implementation:} The input data for the MLP is a history of 150 samples that are flattened into one long array \mbox{$\boldsymbol{x}=\mathbb{F}(\{\boldsymbol{z}_i\}_{i=0}^{150}) \in \mathbb{R}^{6300}$}, \mbox{$\boldsymbol{z}$=$[\boldsymbol{q}, \dot{\boldsymbol{q}}, \boldsymbol{\tau}, \boldsymbol{a}_b, \boldsymbol{\omega}_b] \in \mathbb{R}^{42}$}. For computing model-based results, the input $\boldsymbol{x}$ are transformed into the generalized coordinates: velocities, accelerations, and torques ($\boldsymbol{s}, \dot{\boldsymbol{s}}, \ddot{\boldsymbol{s}}, \boldsymbol{\tau}_s$); with the base linear velocity, base angular acceleration, and joint acceleration estimated by taking derivatives of available measurements. Similar to the contact detection experiment, we re-organize the input data $\boldsymbol{x}$ into a graph structure for our MI-HGNN, with minor modifications to append joint torques to the joint node input and exclude foot inputs. Both MLP and MI-HGNN are trained with a learning rate of $10^{-4}$ and a batch size of 30, using the MSE loss as described in Section~\ref{sec:loss}. The MLP uses a hidden size of 200 with 10 layers, and the MI-HGNN uses a hidden size of 128 with 8 layers, leading to similar parameter sizes as listed in Table \ref{table:modelParams}.

\begin{figure}[t!]
    \centering
    \includegraphics[width=1.0\linewidth]{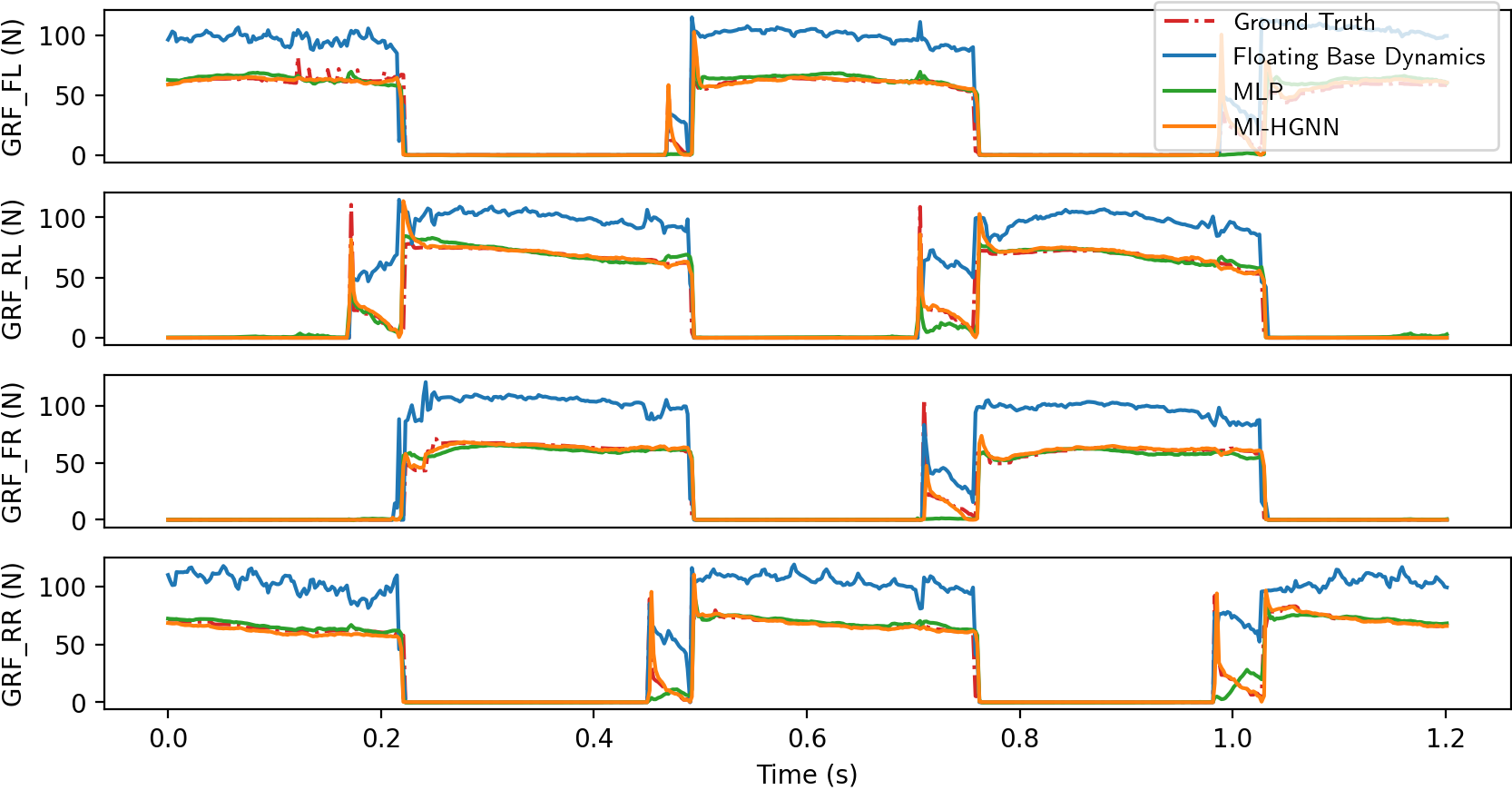}
    \caption{Evaluation of all model types on a sub sequence of the ``Unseen All" test sequence for the GRF estimation task, which includes an unseen friction coefficient ($\dot{x} = 0.5$), unseen speed ($v = 1.0$), and unseen terrain (rough).}
    \label{fig:regression_plot}
\end{figure}

\textbf{Results:} We evaluate each method using RMSE metrics on the test dataset. For the learning-based methods, we train 8 different models and report the average and standard deviation in Table~\ref{table:regression_table}. The MI-HGNN achieves the lowest RMSE on all unseen test sets, and has a lower std. compared with MLP. Fig. \ref{fig:regression_plot} shows the estimated GRFs for each model compared with the ground truth for the most difficult ``Unseen All" test sequence. The FBD model consistently overestimates the GRF, and the MLP exhibits large estimation errors in moments of initial contact with the ground. Overall, the estimated GRFs by our MI-HGNN align better to the ground truth force throughout the test sequence. This experiment shows that our MI-HGNN also achieves superior performance for the GRF estimation task.

\section{CONCLUSION}

In this work, we proposed a Morphology-Informed Heterogeneous Graph Neural Network, that integrates morphological constraints with a graph neural network. Our morphology-aware model imposes constraints in accord with the specific robot morphology to the learning problem, showing superior effectiveness, generalization capability, model efficiency, and sample efficiency for two legged robot contact perception problems. Our future work includes integrating morphological symmetry~\cite{ordonez2023discrete} into our MI-HGNN in a mathematically rigorous manner, extending the current supervised learning to a self-supervised framework using exteroceptive sensors, and implementing the trained model on our robots for real-time estimation. Although in this work, we only applied MI-HGNN on quadruped robots, it can be seamlessly applied to other multi-body dynamical systems including manipulators and humanoid robots.

\section*{ACKNOWLEDGMENT}
We would like to thank F. Xie, Z. Gan and X. Xiong for the valuable discussions, and J. Cheng for data collection.

\bibliographystyle{bib/IEEEtran} %
\bibliography{bib/IEEEabrv, bib/strings-abrv, bib/refs} %

\end{document}